\documentstyle[tree-dvips,proof,ipa,epic,eepic,fullname,11pt]{article}
\title{Type-driven semantic interpretation and feature dependencies in R-LFG}
\author{Mark Johnson\thanks{I would like to thank Mary Dalrymple, 
Ron Kaplan, Dick Oehrle and the anonymous reviewers for this volume,
all of whom were generous with their detailed, insightful 
suggestions.} \\ Brown University}

\newcommand{\Dalrymplerefs}{\cite{Dalrymple95Lamping,Dalrymple96Saraswat,Dalrymple96bSaraswat,Dalrymple96cSaraswat}}

\begingroup
\makeatletter
\newcommand{\tcarray}[1]{{\setbox0=\hbox{$\begin{array}[t]{c} #1 \end{array}$}\raise0.5\ht0\box0}}

\newcommand{\ul}[1]{\underline{#1}}

\newcommand{\semsyn}[2]{\begin{array}[b]{c} #1: \\ #2\end{array}}

\newcommand{\sem}[1]{{\mbox{\it #1}\,'}}
\newcommand{\phon}[1]{{\it #1}}

\newcommand{\gp}[2]{\vtop{\hbox{#1}\hbox{#2}}}

\def\fs#1{$\left[\begin{tabular}{ll}#1\end{tabular}\right]\!\!\!$}
\def\tfs#1{\vbox{\vbox to 0.4em{}\hbox{$\left[\begin{tabular}{ll}#1\end{tabular}\right]\!\!\!$}}}
\def\bfs#1{\vtop{\hbox{$\left[\begin{tabular}{ll}#1\end{tabular}\right]\!\!\!$}}\vbox to 0.4em{}}

\newcommand{\noatt}[1]{\multicolumn{2}{l}{#1}}

\newcommand{\up}{\uparrow\,}
\newcommand{\down}{\downarrow\,}

\newcommand{\eqc}{\mathrel{=_{c}}}

\newcommand{\rlimp}{\mathrel{/}}
\newcommand{\llimp}{\mathrel{\backslash}}
\newcommand{\limp}{\mathrel{-\!\circ}}

\newcommand{\lwith}{\mathbin{\rm \&}}
\newcommand{\lplus}{\oplus}
\newcommand{\lid}{1}

\newcommand{\opt}[1]{\left( #1 \right)}

\newcommand{\cat}[1]{\mbox{#1}}
\newcommand{\lcat}[2]{\begin{array}{c}\cat{#1}\\#2\end{array}}

\renewcommand{\S}{\cat{S}}
\newcommand{\NP}{\cat{NP}}
\newcommand{\VP}{\cat{VP}}
\newcommand{\V}{\cat{V}}

\newcommand{\att}[1]{\ifmmode\mathop{\mbox{\small #1}}\else\mbox{\small #1}\fi}

\newcommand{\SUBJ}{\att{SUBJ}}
\newcommand{\OBJ}{\att{OBJ}}
\newcommand{\XCOMP}{\att{XCOMP}}
\newcommand{\PRED}{\att{PRED}}

\newcommand{\sval}[1]{\mbox{#1}}

\newcommand{\Sandy}{\sval{`Sandy'}}
\newcommand{\professor}{\sval{`professor'}}
\newcommand{\snore}{\sval{`snore$\langle(\up \SUBJ)\rangle$'}}

\newcommand{\CASE}{\att{CASE}}
\newcommand{\NUM}{\att{NUM}}

\newcommand{\GENDER}{\att{GENDER}}
\newcommand{\SG}{\att{SG}}
\newcommand{\PL}{\att{PL}}

\newcommand{\NOM}{\att{NOM}}
\newcommand{\ACC}{\att{ACC}}
\newcommand{\DAT}{\att{DAT}}

\newdimen\widelabel
\def\enumsentence{\@ifnextchar[{\@enumsentence}
{\refstepcounter{equation}\@enumsentence[(\theequation)]}}

\long\def\@enumsentence[#1]#2{\begin{list}{}{%
\advance\leftmargin by\widelabel \advance\labelwidth by \widelabel}
\item[#1] #2
\end{list}}

\newcounter{tempcnt}

\def\@item[#1]{\if@noparitem \@donoparitem
  \else \if@inlabel \indent \par \fi
         \ifhmode \unskip\unskip \par \fi 
         \if@newlist \if@nobreak \@nbitem \else
                        \addpenalty\@beginparpenalty
                        \addvspace\@topsep \addvspace{-\parskip}\fi
           \else \addpenalty\@itempenalty \addvspace\itemsep 
          \fi 
    \global\@inlabeltrue 
\fi
\everypar{\global\@minipagefalse\global\@newlistfalse 
          \if@inlabel\global\@inlabelfalse \hskip -\parindent \box\@labels
             \penalty\z@ \fi
          \everypar{}}\global\@nobreakfalse
\if@noitemarg \@noitemargfalse \if@nmbrlist \refstepcounter{\@listctr}\fi \fi
\setbox\@tempboxa\hbox{\makelabel{#1}}%
\global\setbox\@labels
 \hbox{\unhbox\@labels \hskip \itemindent
       \hskip -\labelwidth \hskip -\labelsep 
       \ifdim \wd\@tempboxa >\labelwidth 
                \box\@tempboxa
          \else \hbox to\labelwidth {\unhbox\@tempboxa}\fi
       \hskip \labelsep}\ignorespaces}

\newcounter{enumsi}

\newdimen\eeindent
\def\@mklab#1{\hfil#1}
\def\enummklab#1{\hfil(\eelabel)\hbox to \eeindent{\hfil#1}}
\def\enummakelabel#1{\enummklab{#1}\global\let\makelabel=\@mklab}
\def\toplabel#1{{\edef\@currentlabel{\p@enums\theequation}\label{#1}}}

\def\eenumsentence{\@ifnextchar[{\@eenumsentence}
{\refstepcounter{equation}\@eenumsentence[\theequation]}}

\long\def\@eenumsentence[#1]#2{\def\eelabel{#1}\let\holdlabel\makelabel%
\begin{list}{\alph{enumsi}.}{\usecounter{enumsi}%
\advance\leftmargin by \eeindent \advance\leftmargin by \widelabel%
\advance\labelwidth by \eeindent \advance\labelwidth by \widelabel%
\let\makelabel=\enummakelabel}
#2
\end{list}\let\makelabel\holdlabel}
\makeatother

\begin{document} \bibliographystyle{fullname} 

\maketitle

\section{Introduction}

\noindent
This paper describes a new formalization of Lexical-Functional Grammar
called R-LFG (where the ``R'' stands for ``Resource-based'').  The
formal details of R-LFG are presented in \namecite{Johnson97}; the
present work concentrates on motivating R-LFG and explaining to
linguists how it differs from the ``classical'' LFG framework
presented in \namecite{Kaplan82}.

This work is largely a reaction to the linear logic semantics for LFG
developed by Dalrymple and colleagues \Dalrymplerefs.  As explained
below, it seems to me that their ``glue language'' approach bears a
partial resemblance to those versions of
Categorial Grammar which exploit the Curry-Howard correspondence
to obtain semantic intepretation
\cite{vanBenthem95}, such as Lambek Categorial Grammar and
its descendants.  A primary goal of this work is to develop a version
of LFG in which this connection is made explicit, and in which
semantic interpretation falls out as a by-product of the Curry-Howard
correspondence rather than needing to be stipulated via semantic
interpretation rules.

Once one has enriched LFG's formal machinery with the linear logic
mechanisms needed for semantic interpretation, it is natural to
ask whether these make any existing components of LFG redundant.
As Dalrymple and her colleagues note, LFG's f-structure completeness
and coherence constraints fall out as a by-product of the linear logic
machinery they propose for semantic interpretation, thus making those
f-structure mechanisms redundant.  Given that linear logic machinery
or something like it is independently needed for semantic
interpretation, it seems reasonable to explore the extent to which it
is capable of handling feature structure constraints as well.

R-LFG represents the extreme position that {\em all} linguistically
required feature structure dependencies can be captured by the
resource-accounting machinery of a linear or similiar logic
independently needed for semantic interpretation.  The goal is to show
that LFG linguistic analyses can be expressed as clearly and
perspicuously using the smaller set of mechanisms of R-LFG as they can
using the much larger set of mechanisms in LFG: if this is the case
then we will have shown that positing these extra f-structure
mechanisms is not linguistically warranted.  One way to show this
would be to present a translation procedure which reduces LFGs to
equivalent R-LFGs, but currently no such procedure is known. Thus we
proceed on a case by case basis, demonstrating that particular LFG
analyses can be expressed at least as well in R-LFG.

R-LFG is also of interest because it proposes a radically different
basis for feature structure interaction.  In ``unification-based''
theories of grammar, feature structures are typically viewed as static
objects, which are the solutions to systems of feature structure
constraints (called {\em f-descriptions} in LFG) 
\cite{Kaplan82,Rounds97,Shieber86}.  
However, linguists often talk informally of ``feature assignment'' and
``feature checking''; notions which cannot be expressed in a pure
unification grammar.  As discussed below, LFG does contain formal
devices which can expresses these notions indirectly, viz., the
non-monotonic devices of existential constraints and constraint
equations.  These have never received an adequate formal description,
despite substantial effort.  On the other hand, the resource oriented
nature of R-LFG provides a direct and natural formalization of the
intuitions behind feature assignment and feature checking.

Because the focus of the work on R-LFG differs from that of the work
of Dalyrmple and her colleagues, the empirical phenomena treated
differ too.  As I understand it, the goal of the ``glue logic'' work
is to provide an account of the syntax-semantics interface which is
compatible with classical LFG syntactic analyses.  The goal of the
R-LFG research is to better understand the relationship between
``resource accounting'' mechanisms and feature structure constraints;
specifically, to determine if the work usually done by feature
structure constraints in LFG might not be done as well or better by
resource mechanisms.  Thus the work in the glue language approach
focusses on semantic phenomena that classical LFG does not account
for, while this paper focusses on syntactic phenomena for which classical LFG
does already describe.

The rest of this paper is structured as follows.  The next section
introduces type-driven semantic interpretation from f-structures,
and the one after that
sketches the architecture of R-LFG and compares it to that of
standard LFG.  The following section introduces the reader to
the idea that features are resources by demonstrating that one
method of describing agreement relationships in standard LFG 
already possesses a resource-oriented character.  The section
following that describes how very simple agreement relationships
can be described in R-LFG, and the final substantive section
shows how \namecite{Andrews82} analysis of Icelandic Quirky Case 
marking can be re-expressed in R-LFG.

\section{Type-driven interpretation from f-structures}

\noindent
This section develops type-driven semantic interpretation
from graph structured resources used in R-LFG, motivating it 
by considering type-driven semantic interpretation from
linearly ordered structures of categories used in Categorial Grammar.

As has often been observed,
the types of semantic objects constrain how they can combine, and
hence the interpretations that can be possibly constructed
from a bag of semantic objects.  For example, suppose the words
\phon{Sandy}\ and \phon{snores}\ are given the semantic interpretations
in (\ref{e:SandySem}) and (\ref{e:snoresSem}) with the types as
shown.
\begin{eqnarray}
 & \sem{Sandy} : e & \label{e:SandySem} \\
 & \lambda x. \sem{snores}(x) : e \limp t & \label{e:snoresSem}
\end{eqnarray}
(The symbol `$\limp$' is the implication symbol of Linear Logic,
so the type $e \limp t$ would be written $e \rightarrow t$ in
a Montagovian notation for types).
Now, there is only one way of combining these semantic objects to form a
saturated proposition of type $t$, namely by applying the semantic
interpretation of the verb \phon{snores}\ to the interpretation
of \phon{Sandy}\ as its argument, so this is the only possible
interpretation of the intransitive clause \phon{Sandy snores}.
This combination can be depicted as a proof (shown in natural deduction
format here), where the two input
semantic forms constitute the assumptions, and the single saturated
proposition produced by the combination constitutes the 
conclusion.\footnote{
The resulting semantic form has been simplified via $\beta$-reduction.}
\[
\infer{\sem{snores}(\sem{Sandy}) : t }
{ \lambda x. \sem{snores}(x) : e \limp t & \sem{Sandy} : e }
\]
It is worth reflecting on what is going on here.
The types alone determine whether a particular way of combining
lexical meanings is possible or not.   The $\lambda$-terms,
which provide the semantic interpretation,
are purely decorative labels: they are completely
determined (up to reduction and
renaming of variables) by the meanings of the lexical inputs
and the structure of the combination. 

The idea that a logic can be used to describe the possible modes
of combination of a collection of objects underlies
the Curry-Howard correspondence, and is at the root of much
recent work in Categorial Grammar \cite{vanBenthem95}.
The formulae of such a logic are the types of the objects being
manipulated, and a proof in this logic corresponds to a particular
way of combining the objects.  The $\lambda$-terms are
decorative labels adorning subproofs that are images
of the structure of the subproof, and play no role in determining
whether a combination is possible or not.

Unfortunately, in more complex sentences semantic type constraints
alone are not sufficiently restrictive to provide just the actually occuring
interpretations.  For example, if the semantic intepretations of
the three words in the sentence
\phon{Sandy likes Kim}\ 
are as given in (\ref{e:SandySem}), (\ref{e:likesSem}) and
(\ref{e:KimSem})
\begin{eqnarray}
 & \lambda y \, \lambda x . \sem{likes}(x,y) : e \limp e \limp t & \label{e:likesSem} \\
 & \sem{Kim} : e & \label{e:KimSem}
\end{eqnarray}
(where `$\limp$' associates to the right)
then besides permitting a combination corresponding to
the available interpretation
\begin{equation} \label{e:goodSem}
\infer{\sem{likes}(\sem{Sandy},\sem{Kim}) : t }
{ \sem{Sandy} : e
& \infer{\lambda x . \sem{likes}(x,\sem{Kim}) : e \limp t}
  { \lambda y \, \lambda x . \sem{likes}(x,y) : e \limp e \limp t
  & \sem{Kim} : e }}
\end{equation}
the semantic type constraints alone also permit an interpretation in which
the subject \sem{Kim}\ and the object \sem{Sandy}\ are exchanged.
\begin{equation} \label{e:badSem}
\infer{\sem{likes}(\sem{Kim},\sem{Sandy}) : t }
{ \infer{\lambda x . \sem{likes}(x,\sem{Sandy}) : e \limp t}
  { \sem{Sandy} : e
  & \lambda y \, \lambda x . \sem{likes}(x,y) : e \limp e \limp t }
& \sem{Kim} : e }
\end{equation}

It is obvious why the unintended interpretation was obtained.  The
semantic types do not reflect any information about the syntactic
structure of the sentence: merely requiring semantic type
compatibility amounts to treating a sentence as a bag of words,
ignoring all other structural relationships between the words.
Clearly this is incorrect for a language like English (as this example
shows).

Standard categorial grammar deals with this problem by refining the structural
sensitivity of the system: the elements manipulated are taken to be a
linearly ordered sequence of categories, rather than just a bag.
Correspondingly, the types are refined to be sensitive
to this additional structural information.
The single implication `$\limp$' used above is specialized
into a rightward-looking implication `$\rlimp$' and a leftward-looking
implication `$\llimp$' respectively.

The types associated with intransitive and transitive verbs
are refined from (\ref{e:snoresSem}) and (\ref{e:likesSem}) to 
(\ref{e:snoresCG}) and (\ref{e:likesCG}), which specify
the directions in which their arguments are to be found.
\begin{eqnarray}
 & \lambda x. \sem{snores}(x) : e \llimp t & \label{e:snoresCG}\\
 & \lambda y \, \lambda x . \sem{likes}(x,y) : (e \llimp t) \rlimp e & \label{e:likesCG})
\end{eqnarray}

This directional sensitivity rules out the unattested combination 
(\ref{e:badSem}), only permitting a combination that corresponds
to the available interpretation.

\[
\infer{\sem{likes}(\sem{Sandy},\sem{Kim}) : t }
{ \sem{Sandy} : e
& \infer{\lambda x . \sem{likes}(x,\sem{Kim}) : e \llimp t}
  { \lambda y \, \lambda x . \sem{likes}(x,y) : (e \llimp t) \rlimp e
  & \sem{Kim} : e }}
\]

Categorial grammarians have developed many insightful linguistic
analyses within this framework.  The treatment of the syntax-semantics
interface within a framework such as Lambek Categorial Grammar and
its descendants is especially appealing: once the lexical types and
modes of syntactic combination are specified, semantic interpretation
comes ``for free'' via the Curry-Howard correspondence between proofs
of type well-formedness and $\lambda$-terms.

However, the focus on linear order in categorial grammar goes against
one of the central intuitions of Lexical-Functional Grammar: that the level
of word order and surface syntactic structure is not an appropriate one
at which to state many cross-linguistic generalizations.  Rather, many
interesting cross-linguistic generalizations are more appropriately
stated at the level of function-argument or f-structure.  

For example, as \namecite{Bresnan82} argues,
the relationship between a verb and its direct object
NP argument may manifest itself cross-linguistically 
in many different surface syntactic 
relationships:
\begin{itemize}
\item it may be indicated by an agreement marker on the verb, or by
\item a case marker on the direct object NP, or by
\item a syntactic configuration, where the object immediately 
      precedes or follows the verb as is appropriate, or by
\item any combination of the above.
\end{itemize}
At the level of function argument structure the cross-linguistic
uniformity of grammatical relation changing operations such as
Passive becomes apparent. 
A central assumption underlying LFG is that a description of
linguistic processes in terms of function-argument relationships
permits simpler and cross-linguistically more uniform accounts
of most linguistic phenomena than would corresponding accounts
in terms of surface syntactic structures.

Thus from an LFG perspective, the appropriate response to the
unattested combination (\ref{e:badSem}) is to make the types
sensitive to function-argument structure rather than
word order directly.  That is, the input to the combinatory
process of semantic interpretation should be f-structures, 
rather than strings of lexical items.

To some extent this is achieved in the work of Dalrymple and her
colleagues.  In their approach, semantic interpretation starts with an
f-structure decorated with formulae from what they call a ``glue
language.''  Semantic interpretation is obtained via a combinatory
process sensitive to function-argument structure.  Moreover, Dalrymple
and colleagues have achieved an impressive empirical coverage using
their glue language approach.

However, the glue language approach seems to suffer from a number of
conceptual drawbacks:
\begin{itemize}
\item The formulae manipulated during the course of a derivation
      are pairs of linear logic terms and standard first-order
      terms connected by the ``glue'' relation `$\leadsto$'.
      While these pairs can be regarded as terms from a (first-order) linear logic,
      this does not seem to be their intended interpretation.
      The term on the right-hand side of the `$\leadsto$' relation obtained
      at the end of the semantic derivation is to be interpreted as
      a classical (higher-order) formula, but no intepretation is provided
      for other pairs appearing in the course of a derivation.
      It would seem to be a weakness of this approach that no
      semantics are provided for these formulae.
\item The semantic combinatory operations in the glue language approach
      are formulated in terms of
      (first-order?) term unification, rather than the function application
      and abstraction operations familiar from model-theoretic semantics.
      It is known from the computational linguistics literature that
      first-order term unification can be used to simulate 
      $\beta$-reduction of $\lambda$-terms in function
      application \cite{Pereira87}, but it is also known that this 
      simulation only 
      approximately captures the properties of function application \cite{Park92}.  
      It would be interesting to see if a system where resources
      have a function-argument structure organization can be made
      to operate with the more standard function application and abstraction
      mechanisms of the $\lambda$-calculus, or if term unification is 
      essential here.\footnote{To the extent to which the glue language
      approach mirrors the account in \namecite{Pereira91a}, it seems
      that unification in the `first-order' formulae on the right-hand
      side of the `$\leadsto$' relation simulates the substitution
      step of $\beta$-reduction.  However, in the abscence of any
      constraints on what constitutes a possible semantic labelling
      in the glue language approach, it is not clear if this property
      will hold of all glue language derivations.}
\item Semantic forms are explicitly constructed in the glue language 
      approach, rather than merely reflecting the structure of
      the proof, as they do in a Lambek Categorial Grammar.
      In principle, the glue language formalism allows semantic
      interpretation rules to be written in which a rule fails to
      apply not because of a type incompatability, but because 
      of unification failure of semantic terms (i.e., terms on the 
      right of the `$\leadsto$' relation).
      Thus these terms need not be restricted to the purely decorative role
      that semantic forms play in Lambek Categorial Grammar, but may determine
      the well-formedness of a proof.\footnote{
In Lambek Categorial Grammar the semantic forms merely record the structure
of the proof, but never act as a filter on proofs.  Thus they do not add
to the complexity of the grammar formalism.}
      Again, it would be interesting to know if this is an
      essential property of semantic interpretation of f-structures,
      or if a system exploiting a Curry-Howard correspondence
      can be developed.
\end{itemize}
Thus the system developed here, R-LFG, is explicitly modelled
on categorial grammars where semantic interpretation is obtained
by a Curry-Howard correspondence.  It differs from them in
that the inputs to the derivational process have the graph
structure of an f-structure, rather than the linear structure
of a string.  

Borrowing the idea that features in feature structures 
can be described by modal operators in a multi-modal language
\cite{Kasper90,Rounds97}, grammatical relations are formalized
as propositional modal operators. 
Returning to the earlier example, the NP \phon{Sandy}\ and
the transitive verb
\phon{likes}\ would be associated with the lexical
entries (\ref{e:SandyR}) and (\ref{e:likesR}).
\begin{eqnarray}
 & \sem{Sandy} : e & \label{e:SandyR} \\
 & \lambda y \, \lambda x . \sem{likes}(x,y) : \OBJ e \limp \SUBJ e \limp t &
  \label{e:likesR}
\end{eqnarray}
(The modal operators `$\SUBJ$', `$\OBJ$', etc., are semantically
vacuous, i.e., always semantically interpreted by identity functions,
and bind more tightly than the implication symbol `$\limp$').  This
entry indicates that the verb \phon{likes}\ first applies to an object
of type $e$ (embedded within the $\OBJ$ grammatical relation),
yielding a function which in turn applies to a subject of type $e$ to
yield a saturated proposition of type $t$.

Assuming that in a transitive clause such as \phon{Sandy likes Kim}\ 
the NP \phon{Sandy}\ can be identified as subject and \phon{Kim}\ 
as object (in English, this occurs by virtue of their c-structure
locations), the following derivation yields the one available
interpretation for this sentence.
\[
\infer{\sem{likes}(\sem{Sandy},\sem{Kim}) : t }
{ \sem{Sandy} : \SUBJ e
& \infer{\lambda x . \sem{likes}(x,\sem{Kim}) : \SUBJ e \limp t}
  { \lambda y \, \lambda x . \sem{likes}(x,y) : \OBJ e \limp \SUBJ e \limp t
  & \sem{Kim} : \OBJ e }}
\]

Following standard treatments of feature structures, re-entrancies are
described by path equations $f_1 \ldots f_m = g_1 \ldots g_n$, which permit
a resource structure $f_1 \ldots f_m \alpha$ to be transformed to
$g_1 \ldots g_n \alpha$.  For example, Subject Raising in LFG is
described in terms of a re-entrancy between the matrix subject position
and the complement's subject position, licensed by a path equation
associated with the Subject Raising verb.  The lexical items in the sentence 
\phon{Sandy seems happy}\ 
would be associated with the lexical entries (\ref{e:SandyR}), (\ref{e:seemsR})
and (\ref{e:happyR}).
\begin{eqnarray}
 & \lambda P . \sem{seems}(P) : \XCOMP t \limp t, \SUBJ = \XCOMP \SUBJ  & \label{e:seemsR} \\
 & \lambda x . \sem{happy}(x) : \SUBJ e \limp t & \label{e:happyR}
\end{eqnarray}
Again, assuming that \phon{Sandy}\ and \phon{happy}\ are identified as
filling the $\SUBJ$ and $\XCOMP$ grammatical functions respectively,
the following deduction shows how the available interpretation for
\phon{Sandy seems happy}\ can be obtained.

\begingroup\small
\renewcommand{\att}[1]{\mathop{\mathstrut\mbox{\footnotesize #1}}}
\[
\infer{\sem{seems}(\sem{happy}(\sem{Sandy})) : t}
{ \semsyn{\lambda P . \sem{seems}(P)}{\XCOMP t \limp t}
& \infer{ \sem{happy}(\sem{Sandy}):\XCOMP t }
  { \infer[*]{ \semsyn{\lambda x . \sem{happy}(x)}{\XCOMP \SUBJ e \limp \XCOMP t} }
    { \semsyn{\lambda x . \sem{happy}(x)}{\XCOMP( \SUBJ e \limp t )} }
  & \infer{ \sem{Sandy} : \XCOMP \SUBJ e }
    { \semsyn{\sem{Sandy}}{\SUBJ e}
    & \SUBJ = \XCOMP \SUBJ
    }
  }
}
\]
\endgroup

The inference labelled `$*$' requires the grammatical relation
$\XCOMP$ to distribute over the implication operator `$\limp$'.

\section{R-LFG: a simplification of LFG}

\noindent
The architectural simplification of R-LFG is best appreciated
when compared with that of standard LFG together with the
linear logic semantics augmentation of Dalrymple and colleagues.
This section starts by sketching the architecture of standard
LFG, and then presents the revised architecture of R-LFG.

\subsection{The architecture of standard LFG}

\begin{figure}
\begin{center}\setlength{\unitlength}{0.00083333in}
\begingroup\makeatletter\ifx\SetFigFont\undefined%
\gdef\SetFigFont#1#2#3#4#5{%
  \reset@font\fontsize{#1}{#2pt}%
  \fontfamily{#3}\fontseries{#4}\fontshape{#5}%
  \selectfont}%
\fi\endgroup%
{\renewcommand{\dashlinestretch}{30}
\begin{picture}(6085,3639)(0,-10)
\path(3312,2412)(3312,2112)
\path(2412,2112)(4212,2112)(4212,1812)
	(2412,1812)(2412,2112)
\put(3237,2187){\makebox(0,0)[rb]{\smash{{{\SetFigFont{10}{12.0}{\rmdefault}{\mddefault}{\itdefault}defines}}}}}
\put(3312,1887){\makebox(0,0)[b]{\smash{{{\SetFigFont{10}{12.0}{\familydefault}{\mddefault}{\updefault}minimal f-structures}}}}}
\path(3762,312)(5862,312)(5862,12)
	(3762,12)(3762,312)
\path(4812,612)(4812,312)
\path(3762,912)(5862,912)(5862,612)
	(3762,612)(3762,912)
\put(4812,87){\makebox(0,0)[b]{\smash{{{\SetFigFont{10}{12.0}{\familydefault}{\mddefault}{\updefault}semantic interpretation}}}}}
\put(4812,687){\makebox(0,0)[b]{\smash{{{\SetFigFont{10}{12.0}{\familydefault}{\mddefault}{\updefault}glue language formula}}}}}
\put(4962,1062){\makebox(0,0)[lb]{\smash{{{\SetFigFont{10}{12.0}{\rmdefault}{\mddefault}{\itdefault}semantic mapping}}}}}
\put(4962,387){\makebox(0,0)[lb]{\smash{{{\SetFigFont{10}{12.0}{\rmdefault}{\mddefault}{\itdefault}linear logic proof}}}}}
\path(2262,3012)(2262,2862)(1212,2862)(1212,2712)
\path(2262,2862)(3312,2862)(3312,2712)
\path(2712,2712)(3912,2712)(3912,2412)
	(2712,2412)(2712,2712)
\dashline{60.000}(1812,2562)(2712,2562)
\path(1212,3162)(1212,3012)(3237,3012)(3237,3162)
\path(312,2112)(2112,2112)(2112,1812)
	(312,1812)(312,2112)
\path(1212,2412)(1212,2112)
\path(612,2712)(1812,2712)(1812,2412)
	(612,2412)(612,2712)
\path(2787,3462)(3837,3462)(3837,3162)
	(2787,3162)(2787,3462)
\path(462,3462)(1962,3462)(1962,3162)
	(462,3162)(462,3462)
\path(3312,1812)(3312,1512)
\path(2412,1512)(4212,1512)(4212,1212)
	(2412,1212)(2412,1512)
\path(3312,1212)(3312,1062)(4812,1062)
\path(3387,3162)(3387,3012)(4812,3012)(4812,912)
\put(3312,2487){\makebox(0,0)[b]{\smash{{{\SetFigFont{10}{12.0}{\familydefault}{\mddefault}{\updefault}f-description}}}}}
\put(1212,1887){\makebox(0,0)[b]{\smash{{{\SetFigFont{10}{12.0}{\familydefault}{\mddefault}{\updefault}phonological form}}}}}
\put(1137,2187){\makebox(0,0)[rb]{\smash{{{\SetFigFont{10}{12.0}{\rmdefault}{\mddefault}{\itdefault}yields}}}}}
\put(1212,2487){\makebox(0,0)[b]{\smash{{{\SetFigFont{10}{12.0}{\familydefault}{\mddefault}{\updefault}c-structure}}}}}
\put(3312,3237){\makebox(0,0)[b]{\smash{{{\SetFigFont{10}{12.0}{\familydefault}{\mddefault}{\updefault}Lexicon}}}}}
\put(1212,3237){\makebox(0,0)[b]{\smash{{{\SetFigFont{10}{12.0}{\familydefault}{\mddefault}{\updefault}Syntactic Rules}}}}}
\put(1062,2862){\makebox(0,0)[rb]{\smash{{{\SetFigFont{10}{12.0}{\rmdefault}{\mddefault}{\itdefault}generates}}}}}
\put(3237,1587){\makebox(0,0)[rb]{\smash{{{\SetFigFont{10}{12.0}{\rmdefault}{\mddefault}{\itdefault}constraint filter}}}}}
\put(3312,1287){\makebox(0,0)[b]{\smash{{{\SetFigFont{10}{12.0}{\familydefault}{\mddefault}{\updefault}minimal f-structures}}}}}
\dottedline{45}(12,3612)(4512,3612)(4512,1137)
	(12,1137)(12,3612)
\end{picture}
}\end{center}
\caption{The architecture of standard LFG.  The linear logic semantics
component is shown outside the dotted box. \label{f:lfg1a}}
\end{figure}
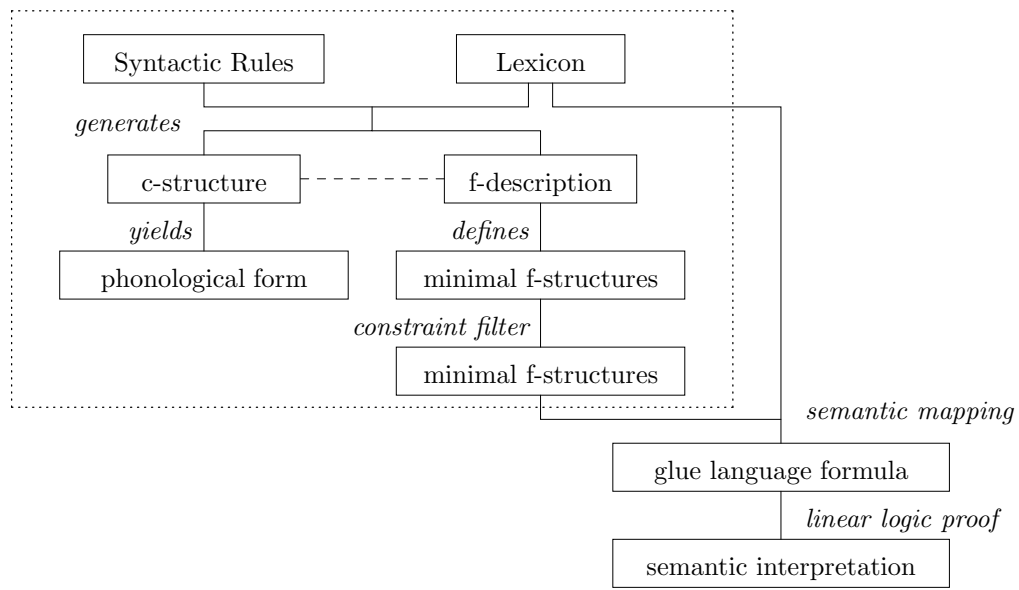

\noindent
Figure~\ref{f:lfg1a} shows the architecture of ``standard'' LFG.
The components of LFG as presented by \namecite{Kaplan82} are
shown inside the dotted box in this figure, and the linear logic
machinery for semantic interpretation posited by Dalrymple and colleagues
is depicted outside this box.

In LFG, a syntactic description of an utterance is taken to be a pair
constiting of a c-structure and an f-structure.\footnote{There are
proposals for additional structures, which for simplicity are ignored
here.}  The yield of the c-structure tree determines the phonological
form of the sentence it describes.

The c-structure/f-structure pairs generated by an LFG are determined
by the following procedure.  The syntactic rules and lexical entries
of an LFG together generate a set of c-structure trees, each of which
is paired with a formula called an f-description which identifies
which (if any) f-structures this c-structure can be paired with.
The f-descriptions are boolean combinations of equations.  These
equations come in two kinds: {\em defining} and {\em constraining}
equations.  

The simplest account of the relationship between f-descriptions and
the f-structures they describe seems to be procedural, following
\namecite{Kaplan82}.  First, the f-description is expanded into
Disjunctive Normal Form (DNF) and the f-structure solution to each
conjunct is determined as follows. The constraining equations are
temporarily ignored (i.e., replaced with {\em true}) and if the
resulting formula is satisfiable and has a unique minimal satisfying
f-structure, that f-structure is a candidate solution to the conjunct.
This candidate solution is a (true) solution to the conjunct just in
case it also satisfies the formula obtained by replacing each
constraining equation in the conjunct with corresponding defining
equations.  The set of solutions to an f-description is the union of
the set of solutions to each conjunct of its DNF, so the f-description
determines a finite number of f-structures.\footnote{
\renewcommand{\att}[1]{\mathop{\mbox{\tiny #1}}}
To appreciate some of the difficulties in giving a declarative
treatment of LFG's constraint equations, 
consider a treatment of Case marking in
which subject \NP s are optionally assigned a nominative Case
feature $\NOM$, such as the \namecite{Andrews82} analysis of Icelandic
quirky case marking discussed in section~\ref{s:Icelandic}, using the
following LFG syntactic rule.
\[
\S \, \rightarrow \, \lcat{NP}{(\up \SUBJ) = \down\\((\up \SUBJ \CASE) = \NOM)} \, \lcat{VP}{\up = \down}
\]
The parentheses surrounding the lower equation annotating the \NP\ indicates
that this defining equation is optional, reflecting the fact that the
subject \NP\ is only optionally assigned nominative case (as it may
be assigned a `quirky' non-nominative case by the verb, as explained
below).  This annotation presumably abbreviates
the following disjunction:
\[
 (\up \SUBJ \CASE) = \NOM \vee \mbox{\it true}
\]
Clearly replacing this disjunction with \mbox{\it true} does not change
the set of minimal models for any f-description which contains it,
so the equation itself has no effect on the minimal models, and hence
cannot result in the satisfaction of any constraint equations.  Clearly
this is not the intended interpretation: the ``purpose'' of this equation
is to provide a Case feature to satisfy the requirements of the subject
$\NP$.

\namecite{Kaplan82} do not discuss disjunction, but it appears they
intend disjunctions to be interpreted as an abbreviatory convention,
i.e., that their process applies only to individual conjunctions after
expansion to a Disjunctive Normal Form (DNF).  Thus their treatment,
while not falling foul of the problem just noted, involves a rather
curious mixture of proof-theoretic devices (e.g., DNF expansion) and
model-theoretic devices (e.g., focussing on minimal models).}

Dalrymple et.\ al.\ use these f-structures as the input to their
semantic interpretation procedure.  Certain elements in an f-structure
are associated with formulae in a {\em glue language}, which is an
amalgam of linear logic and classical first-order logic, in effect
mapping each f-structure into a formula of the glue language.  For
semantic interpretation to succeed this glue language formula must
derive a term with the type of a saturated proposition: the argument
of this term is the semantic interpretation of the sentence.

\subsection{The architecture of R-LFG}

\begin{figure}
\begin{center}\setlength{\unitlength}{0.00083333in}
\begingroup\makeatletter\ifx\SetFigFont\undefined%
\gdef\SetFigFont#1#2#3#4#5{%
  \reset@font\fontsize{#1}{#2pt}%
  \fontfamily{#3}\fontseries{#4}\fontshape{#5}%
  \selectfont}%
\fi\endgroup%
{\renewcommand{\dashlinestretch}{30}
\begin{picture}(4374,2364)(0,-10)
\path(1062,2037)(1062,1887)(2862,1887)(2862,2037)
\path(312,2337)(1812,2337)(1812,2037)
	(312,2037)(312,2337)
\path(1962,1887)(1962,1737)
\path(1362,1737)(2562,1737)(2562,1437)
	(1362,1437)(1362,1737)
\path(12,1137)(1662,1137)(1662,837)
	(12,837)(12,1137)
\path(2562,1137)(3762,1137)(3762,837)
	(2562,837)(2562,1137)
\path(1962,1437)(1962,1287)
\path(3162,837)(3162,537)
\path(2412,2337)(3462,2337)(3462,2037)
	(2412,2037)(2412,2337)
\path(3012,2037)(3012,1887)(3237,1887)(3237,1137)
\path(762,1137)(762,1287)(3087,1287)(3087,1137)
\path(1962,537)(4362,537)(4362,12)
	(1962,12)(1962,537)
\put(1062,2112){\makebox(0,0)[b]{\smash{{{\SetFigFont{10}{12.0}{\familydefault}{\mddefault}{\updefault}Syntactic Rules}}}}}
\put(1962,1512){\makebox(0,0)[b]{\smash{{{\SetFigFont{10}{12.0}{\familydefault}{\mddefault}{\updefault}c-structure}}}}}
\put(837,912){\makebox(0,0)[b]{\smash{{{\SetFigFont{10}{12.0}{\familydefault}{\mddefault}{\updefault}phonological form}}}}}
\put(3162,912){\makebox(0,0)[b]{\smash{{{\SetFigFont{10}{12.0}{\familydefault}{\mddefault}{\updefault}f-term}}}}}
\put(3162,312){\makebox(0,0)[b]{\smash{{{\SetFigFont{10}{12.0}{\familydefault}{\mddefault}{\updefault}type well-formedness proof}}}}}
\put(3312,612){\makebox(0,0)[lb]{\smash{{{\SetFigFont{10}{12.0}{\rmdefault}{\mddefault}{\itdefault}proof}}}}}
\put(912,1812){\makebox(0,0)[rb]{\smash{{{\SetFigFont{10}{12.0}{\rmdefault}{\mddefault}{\itdefault}generates}}}}}
\put(2937,2112){\makebox(0,0)[b]{\smash{{{\SetFigFont{10}{12.0}{\familydefault}{\mddefault}{\updefault}Lexicon}}}}}
\put(687,1287){\makebox(0,0)[rb]{\smash{{{\SetFigFont{10}{12.0}{\rmdefault}{\mddefault}{\itdefault}labelling}}}}}
\put(3162,87){\makebox(0,0)[b]{\smash{{{\SetFigFont{10}{12.0}{\familydefault}{\mddefault}{\updefault}= semantic interpretation}}}}}
\end{picture}
}\end{center}
\caption{The architecture of R-LFG. \label{f:lfg2}}
\end{figure}
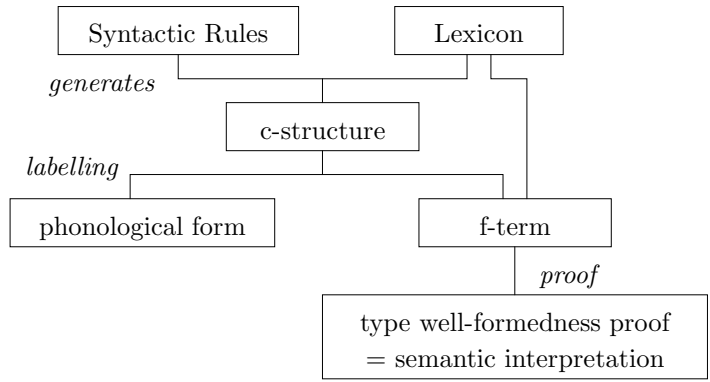

\noindent
The architecture of R-LFG is depicted in Figure~\ref{f:lfg2}.
The most striking difference between LFG and R-LFG is that
R-LFG does not contain an independent level of f-structure
representation, since the same mechanisms used for semantic
interpretation are also used to account for syntactic feature
dependencies.  Given that it is a simpler architecture,
it should be preferred on grounds of parsimony.

The lexical entries and syntactic rules of R-LFG generate
c-structure/f-term pairs in the same way that they generate
c-structure/f-description pairs in LFG.  In LFG several steps are
required to obtain the f-structures that serve as the input to
semantic interpretation from the f-descriptions.  However, in R-LFG
the f-term serves as the input to semantic interpretation directly.
Thus in R-LFG the linguistic effects of f-structure constraints
must be obtained by other means, viz., the same logical mechanisms
used for semantic interpretation.

As explained below, these logical mechanisms enforce a {\em resource
accounting} which ensures that every predicate combines with an
appropriate number of arguments and that every non-root semantic unit 
appears as the argument of some predicate.  The semantic interpretation
itself is determined by the pattern of predicate-argument combination
via a Curry-Howard correspondence, as explained 
in more detail in \namecite{Johnson97}.

This same resource accounting mechanism is also used to describe
feature dependencies.  Purely syntactic features with no semantic
content differ from semantically interpreted elements only in that
they are semantically vacuous, i.e., given trivial interpretations
which are systematically ignored by any functors which take them as
arguments.

The resource logic used here differs considerably from the glue
language used by Dalrymple et.\ al.  That language includes
first-order terms with equality, which can be used to encode feature
structure unification in the manner of e.g., Definite Clause Grammars,
and hence directly
simulate f-structure attribute-value constraints
(see \namecite{Shieber86} for a description of the relationship
between the first-order terms of Definite Clause Grammars and
attribute-value ``unification'' grammars).  While this
would provide a straightforward way to encode f-structure
constraints in the glue language, it is not clear that such
an approach would constitute a real simplification of LFG,
rather than just a reshuffling of its complexity.

For this reason, R-LFG uses a much simpler resource logic than the
glue language of Dalrymple et.\ al.  Inspired by recent work in
Categorial Grammar such as \namecite{Morrill94} the resource logic is based on a
propositional modal logic, which encodes the types of the semantic
objects being manipulated, and the semantic interpretation itself is
provided by a Curry-Howard correspondence between proofs and
$\lambda$-terms \cite{Girard89}.  As \namecite{vanBenthem95}
demonstrates, a wide variety of substructural logics possess a
Curry-Howard correspondence, so the requirement that semantic
interpretation is obtained in this way does not identify a particular
logic.  Rather, the precise logic used should be chosen to best fit
the linguistic phenomena described by the theory.  
\namecite{Moortgat:Handbook} develops the
theory of propositional multimodal logics used here.  

\section{Describing agreement relationships with LFG}

\noindent
This section argues that Lexical-Functional grammarians typically
use the formal devices of LFG to manipulate features as resources
that are assigned and checked.
It introduces two methods often used for describing agreement
relationships in LFGs.  It turns out that one method, which crucially
relies on ``constraining equations'', can be viewed as describing
agreement in terms of resource dependencies.  Thus resource-based
accounts of agreement are not a new innovation of R-LFG, but are
already a familiar part of LFG.  The principal claim behind R-LFG
is that {\em all} linguistic dependencies can be expressed in this
manner, and that the explicit resource-orientation of R-LFG simplifies
and clarifies the nature of the linguistic dependencies concerned.

As explained in more detail in \namecite{Kaplan82},
LFG's f-descriptions contain two different kinds of equations.  A
defining equation instantiates the value of an attribute, while a
constraining equation checks that a value is instantiated
by a defining equation elsewhere in the f-description.
The linguistic dependencies involved in simple agreement can
be described using defining equations alone, or by using
a mixture of defining and constraining equations.
This latter method has a natural resource interpretation.

To keep things clear, the two methods for describing agreement
relationships are explained using the same examples (\ref{e:agr1}).
\eenumsentence{\label{e:agr1} 
 \item Sandy snores. \label{e:agr1a}
 \item Professors snore.}  
Both methods of describing agreement relationships require that the
agreeing items (in (\ref{e:agr1a}), \phon{Sandy} and \phon{snores})
are capable of constraining the value of the same f-structure element;
this is usually achieved by defining equations associated with
syntactic rules.  The agreeing items both impose constraints on the value
of that shared f-structure element, thus ensuring that only compatible items
can appear simultaneously in a syntactic structure.

\subsection{Agreement using defining equations alone}

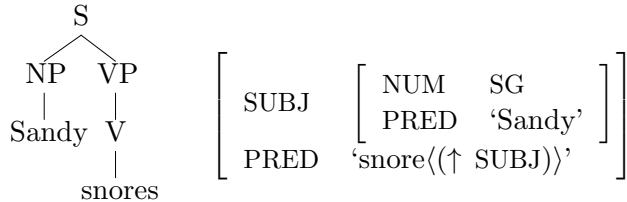
\begin{figure}
\begin{center}
\mbox{\lower3em\hbox{
\begin{picture}(54,78)(0,-78)
%
\put(24,-8){S}
\drawline(27,-12)(13,-22)
\put(6,-30){NP}
\drawline(13,-34)(13,-44)
\put(0,-52){Sandy}
\drawline(27,-12)(40,-22)
\put(33,-30){VP}
\drawline(40,-34)(40,-44)
\put(36,-52){V}
\drawline(40,-56)(40,-66)
\put(27,-74){snores}
\end{picture}}}
\hspace{0.2in}
\fs{ \SUBJ & \tfs{ 
                   \NUM  & \SG \\
                   \PRED & \Sandy } \\
     \PRED & \snore }
\end{center}
\caption{\label{f:cf1} 
  The c-structure and f-structure for \phon{Sandy snores} generated
  by the fragment (\ref{e:gr1a}--\ref{e:gr1e}).}
\end{figure}

\noindent
In this method, both agreeing items constrain the shared f-structure
element using defining equations.  For example, the 
grammar fragment in (\ref{e:gr1a}--\ref{e:gr1e})
generates exactly the two sentences in (\ref{e:agr1}).  The
c-structure and f-structure generated by this fragment for
(\ref{e:agr1a}) is depicted in Figure~\ref{f:cf1}.

\begin{eqnarray}
 \label{e:gr1a} \phon{Sandy} & \NP &
 \begin{array}[t]{l} (\up \PRED) = \Sandy\\ (\up \NUM) = \SG \end{array} \\
 \label{e:gr1b} \phon{Professors} & \NP &
 \begin{array}[t]{l} (\up \PRED) = \professor\\ (\up \NUM) = \PL \end{array} \\
 \label{e:gr1c} \phon{snores} & \VP &
 \begin{array}[t]{l} \up \PRED) = \snore\\ \underline{(\up \SUBJ\;\NUM) = \SG}
 \end{array}\\
 \label{e:gr1d} \phon{snore} & \VP &
 \begin{array}[t]{l} \up \PRED) = \snore\\ \underline{(\up \SUBJ\;\NUM) = \PL}
 \end{array}
\end{eqnarray}
\begin{equation} \label{e:gr1e}
 \S \; \longrightarrow \;
  \begin{array}[t]{c} \NP \\ (\up\SUBJ) = \down \end{array} 
  \begin{array}[t]{c} \VP \\ \up = \down \end{array}
\end{equation}

The lexical entries for subject $\NP$s require that the value of their
$\NUM$ attribute is $\SG$ or $\PL$ as appropriate.  In addition, the
underlined equation in each verb's lexical entry also requires that
this value is appropriate for the verb's inflection.  If the subject
and the verb require different values for this f-structure element (as
in the ungrammatical \phon{*Professors snores}), the corresponding
f-description will require this element to be equal to two different
values (e.g., $\SG$ and $\PL$).  However, the well-formedness
conditions on f-structures do not permit this
\cite{Kaplan82,Johnson95b} so the f-descriptions associated with such
sentences are inconsistent, and the sentences themselves are correctly
predicted to be ungrammatical.  

Thus this method functions by arranging for ungrammatical sentences to
be associated with an inconsistent f-description.  This observation is
in fact quite general: if all grammatical relationships are described
using defining equations (i.e., if we restrict
attention to the monotonic constraints) then the only way such
an equation can have a grammatical ``effect'' is by being inconsistent
with other equations, i.e., by ``causing'' ungrammaticality.

More precisely, suppose we identify a subset of the elements of an
f-structure as follows.  The {\em semantically interpreted elements}
are those which serve as the input to the semantic interpretation
procedure (in the framework of Dalrymple et.\ al.\ these elements
are associated with glue language formulae at some stage during the
interpretation process).  The idea is the semantically uninterpreted
elements can be deleted from an f-structure without changing its
semantic interpretation.  In a typical LFG, the values of attributes
such as $\PRED$, $\SUBJ$, $\OBJ$, etc., are semantically interpreted,
while the values of $\CASE$ and $\GENDER$ (in a grammatical gender
language) are not semantically interpreted.

Now consider a ``pure unification'' grammar without non-monotonic
devices such as ``constraining equations'', e.g., in which all
equations are defining equations, such as the PATR grammars of
\namecite{Shieber86}.  These are grammars in which all linguistic
relationships are expressed with defining equations.  It is
possible to show that in such a grammar, if an equation which equates
only non-semantic values is not inconsistent with other equations on
some input, then deleting it from the grammar does not affect the
language generated or the interpretations assigned.  (A similiar
observation holds in monotonic grammars such as HSPG).

This means that if all grammatical relationships are described using
defining equations, a
nonsemantic feature defining equation only has an effect on
the language generated if somewhere else in the
grammar there are defining equations that are inconsistent with this
one.
For example, there is no point in adding a defining equation that
introduces an attribute that does not appear elsewhere in the grammar,
such as
\begin{equation} \label{e:romance}
     (\up \att{HISTORICAL-ORIGIN}) = \att{ROMANCE}
\end{equation}
unless other defining equations that can possibly be inconsistent
with it are also introduced.  But in order to be inconsistent
with (\ref{e:romance}) these other equations must require the
attribute's value to be {\em different} to the value specified
in the former equation, e.g.,
\[
     (\up \att{HISTORICAL-ORIGIN}) = \att{GERMANIC}.
\]

Thus with defining equations alone, different grammatical properties
are based on feature {\em oppositions} or constrasts.  The formal
machinery of these monotonic ``pure unification'' grammars does not
completely support non-constrastive or ``privative'' feature values.

Indeed, f-structures seem to have been specifically designed to enable
systems of defining equations to be inconsistent.  For example, if we
removed either the ``functionality'' axiom (which requires attributes
to be single-valued) or the ``constant-constant'' clash axiom (which
specifies that distinct constants denote distinct f-structure
elements) from the formal definition of f-structures, then
f-descriptions such as
\[
	(f\; \CASE) = \ACC, (f\;\CASE) = \DAT
\]
would not be inconsistent.  R-LFG does not possess either the
functionality axiom or the constant-constant clash axiom,
and hence it does permit a single constituent to bear two
such distinct features, so long as both are checked or consumed
as described below.

\subsection{Agreement using defining and constraining equations}

\noindent
Writers of LFGs typically employ constraining equations in order
to describe asymmetric linguistic relationships.  The subject-verb
agreement examples (\ref{e:agr1}) would be described using this method
by replacing the lexical entries (\ref{e:gr1c}--\ref{e:gr1d}) with
the following. 
\begin{eqnarray}
 \label{e:gr2c} \phon{snores} & \VP &
 \begin{array}[t]{l} \up \PRED) = \snore\\ \underline{(\up \SUBJ\;\NUM) \eqc \SG}
 \end{array}\\
 \label{e:gr2d} \phon{snore} & \VP &
 \begin{array}[t]{l} \up \PRED) = \snore\\ \underline{(\up \SUBJ\;\NUM) \eqc \PL}
 \end{array}
\end{eqnarray}
These entries differ from the previous ones in that the underlined
defining equations have been replaced with constraining equations.

While these two fragments both generate the same language in this
case, in general the two methods for describing agreement behave quite
differently.  For example, if an \NP's f-description contains the
constraint equation
\begin{equation} \label{e:constraint}
  (\up \CASE) \eqc \ACC 
\end{equation}
then this \NP\ must be independently ``assigned'' a value for the Case
feature in order for the f-structure to be well-formed.

This method behaves quite differently to the method that only uses
defining equations.  It does not rely on feature oppositions in the
same way that the defining equation method does.  For example, the
constraint equation (\ref{e:constraint}) requiring that the $\NP$
receive an $\ACC$ case value does not rely on the existence of other
Case values besides $\ACC$; it functions just as well if $\ACC$ is the
only Case value used in the grammar.  That is, while a defining
equation ensures that an attribute has one value rather than another,
a constraining equation ensures in addition that the feature has in
fact been given a value independently.  Thus this method more fully
supports privative features than the defining equation method does.

Further, the constraining equation method does not rely on the
functionality axiom or the constant-constant clash axioms in the same
way that the defining equation method does.  For example, even if
the functionality requirement on f-structures were relaxed so that the
defining equations in the f-description for (\ref{e:agr1a}) could have
the second minimal f-structure solution depicted in
Figure~\ref{f:nonfunct} besides the one depicted in
Figure~\ref{f:cf1}, that f-structure would fail to satisfy the
constraining equation expressing subject-verb agreement, and
so would be ill-formed for independent reasons.

\begin{figure}
\begin{center}
\fs{\SUBJ & \tfs{ \PRED & \Sandy \\
                  \NUM  & \SG } \\
    \SUBJ & \tfs{ \NUM   & $\kern-1.5ex\eqc$ \SG } \\
    \PRED & \snore }
\end{center}
\caption{A alternative minimal f-structure solution to the
 f-description for (\protect\ref{e:agr1a}) obtained by relaxing
 the functionality requirements on f-structures.  Note that this
 f-structure never the less does not satisfy the constraining
 equations expressing subject-verb agreement because the constraint
 equation embedded in the lower
 \SUBJ is not satisfied. \label{f:nonfunct}}
\end{figure}


In fact, feature structures in R-LFG behave very much in this way.
While attributes are permitted to be single-valued, no feature
structure axiom forces them to be so.  But since grammatical relationships
are described in a way very similiar to the constraining equation method,
in general the grammatical requirements of predicates will require
that attributes are single-valued.  However, `single-valuedness' is not
built into the R-LFG formalism the way it is in standard LFG, opening
the possibility of analyses which require multiple instantiations of
the same grammatical relation within a single clause.

\subsection{Resource management in LFG}

\noindent
The constraining equation method of describing agreement relationships
can be described in terms of {\em resources}, where the resource is
the feature value of the shared f-structure entity.  Each such feature
value is {\em produced} by {\em one or more} defining equations,
and is {\em consumed} by {\em zero or more} constraining equations.
This pattern of resource management is formalized by Intuitionistic
Logic.

Interestingly, the special properties LFG endows the values of
$\PRED$ attributes with provides them with special resource
management properties also.  The values of $\PRED$ attributes
must be {\em produced} by {\em exactly one} argument, and must be 
{\em consumed} by {\em one or more} predicates.  The logic
LPC developed by \namecite{vanBenthem95}
formalizes this resource management.

Thus LFG already incorporates a number of mechanisms which can be seen
as performing resource management.  R-LFG attempts to describe all
syntactic relationships in terms of such resource management.  Identifying
the appropriate resource management mode for a particular grammatical
relationship is a key step in developing its R-LFG description.

It is interesting that Multiplicative Linear Logic (MLL) enforces a different resource
managment regime than either Intuitionistic Logic or LPC
(MLL requires each resource to be produced exactly once and
consumed exactly once), although it can simulate other modes by
means of its exponential operators \cite{Girard95}.
For more discussion of appropriate resource managment in LFG,
particularly controlled applications of Contraction, see \namecite{Johnson97}.

\section{Resource accounting in R-LFG}

\noindent
\namecite{Johnson97} formally defines R-LFG's f-terms and presents a
Gentzen sequent calculus that describes the resource management 
relationships between features.  It also presents labelled deduction
systems for describing the mappings from c-structures to f-terms, and
semantic intepretation from f-terms.
That paper should be consulted for the technical details of R-LFG;
this section presents that material in an informal and hopefully
more accessible manner.

An f-formula is an expression that indicates the type of a constituent,
or more generally, a single resource.
The semantic type of a constituent can be determined from its f-formula,
but just as in the categorial grammar example above, f-formulae also
specify additional syntactic constraints.   

Following \namecite{Morrill94}, we distinguish semantically contentful
types from semantically impotent types.  The basic semantically
contentful types $e$, $t$, etc., are f-formulae (these are the types
of individuals and truth values respectively), as are the basic
semantically impotent types $\NOM$, $\ACC$, etc., (which are
interpreted by constants, and whose value is systematically ignored by
any function that takes them as an argument).  The full set of
f-formulae used here are obtained by closing these under the following operations.

\begin{description}
\item  If $\varphi$ is an f-formula then $f\, \varphi$ is also an f-formula,
       where $f$ is an attribute; it denotes the result of embedding $\varphi$
       under the attribute $f$.
\item  If $\varphi_1$, $\varphi_2$ are f-formulae then $\varphi_1 \limp \varphi_2$
       is also an f-formula; it is a linear implication which consumes
       $\varphi_1$ to produce $\varphi_2$.
\end{description}

To formulate larger grammars it would be worthwhile introducing additional
Linear Logic connectives.  For example, the additive connective
`$\lwith$' provides disjunction of features, while the additive
connective `$\lplus$' can be used to express the ``overspecified''
features required by the \namecite{Bayer95} analysis of feature
distributivity in coordination.  Indeed, it is straightforward
to translate these analyses into R-LFG. \namecite{Johnson97} shows how
optionality can be expressed using the additive connective
`$\lwith$' and the additive identity `$\lid$'.

The relationship between f-formulae and the more usual types of
model-theoretic semantics is given by the mapping $(\cdot)^\natural$,
which maps f-formulae to standard model-theoretic types.
In this mapping $\emptyset$ is a new type constant interpreted
by a single element domain that is used to interpret 
semantically impotent f-formulae.
\begin{description}
\item[$(\varphi)^\natural =  \varphi$] if $\varphi$ is a semantically contentful
 basic type,
\item[$(\varphi)^\natural = \emptyset$] if $\varphi$ is a semantically impotent
 basic type,
\item[$(f\, \varphi)^\natural = (\varphi)^\natural$] where $f$ is an attribute, and
\item[$(\varphi_1 \limp \varphi_2)^\natural = (\varphi_2)^\natural$] if 
      $(\varphi_1)^\natural = \emptyset$, and 
      $(\varphi_1)^\natural \rightarrow (\varphi_2)^\natural$ otherwise.
\end{description}
For example, the natural type of an f-formula for an $\NP$ requiring
a nominative case marking is
$
(\NOM \limp e)^\natural \; = \; e.
$
In general, it is required that any $\lambda$-term labelling an
f-formulae $\varphi$ (i.e., giving the constituent's semantic interpretation)
be of type $(\varphi)^\natural$.  (Semantically impotent f-formulae are
not labelled with $\lambda$-terms, as they have no natural semantic
interpretation).

F-formulae are the building blocks of f-terms.
Informally, an f-term is a graph-structured configuration 
of one or more constituents, or more generally, resources.
F-formulae are f-terms, and if $\alpha, \alpha_1, \ldots, \alpha_n$
are f-terms then:
\begin{description}
\item[$\alpha_1, \ldots, \alpha_n$] is the {\em multiset} of 
 resource structures $\{ \alpha_1, \ldots, \alpha_n \}$ (order is 
 unimportant in a multiset,
 but the number of times an element appears is important),
\item[$f\;\alpha$] is the result of {\em embedding} the structure 
 $\alpha$ under the attribute $f$,\footnote{
\namecite{Johnson97} follows \namecite{Moortgat:Handbook} in 
introducing a separate punctuation symbol to distinguish modal structures in
f-terms from modal operators in f-formulae, but here we rely on
context to distinguish these two usages.}
\item[$f_1 \ldots f_m = g_1 \ldots g_n$] is a path equation which 
 {\em restructures} an f-term
 by moving a resource structure embedded under the sequence of
 attributes $f_1 \ldots f_m$ so that it is located under the sequence
 of attributes $g_1 \ldots g_n$, and
\item[$\opt{\alpha}$] is an {\em optional} occurence of the structure
 $\alpha$.
\end{description}

An f-term describes a graph structure of constituents, or more
generally, resources.  The f-term associated with a sentence is
required to simplify to a single resource of type $t$ in order for the
sentence to be grammatical.  (This single requirement subsumes both
the requirement that the f-description be satisfiable and the
requirement that the Linear Logic glue formula simplify to an
expression of type $t$ in standard LFG).  An f-term simplifies by
applying linear implications, restructuring using path equations,
distributing attributes over multisets and implications, and either
deleting optional elements or replacing them with their non-optional
counterpart.

Attributes are permitted, but not required, to distribute and factor 
over multisets.
That is, the following bi-implication holds, where $f$ is an attribute
and $\alpha_1$ and $\alpha_2$ are f-terms:
\[
 f(\alpha_1 , \alpha_2) \;\Leftrightarrow\; (f\,\alpha_1), (f\,\alpha_2).
\]
Unlike LFG, R-LFG does not require that attributes are single-valued,
nor does it enforce a constant-constant clash.  Every f-term is
``satisfiable'' in that it represents some configuration of resources;
grammaticality is determined by whether those resources can combine
to produce a single element of type $t$ (the type of a saturated
proposition).

\subsection{Nominative Case marking in English}

\noindent
A simple R-LFG fragment which describes structural nominative case
assignment to subject $\NP$s is presented below.  The lexical entry
for the nominative Case marked subject NP \phon{Sandy} in 
(\ref{e:she}) requires it to consume a
$\NOM$ case resource in order to produce a resource of type $e$, and
the lexical entry for the verb \phon{snores} in (\ref{e:snores})
requires it to consume a resource of type $e$ embedded within a
$\SUBJ$ attribute in order to produce a resource of type $t$.  

The
syntactic rule (\ref{e:SNPVP}) specifies how the f-terms associated
with the $\NP$ and $\VP$ (referred to by the meta-variable `$\down$' just
as in LFG) are to be combined to produce the f-term for the $\S$.  
In
this case, a multiset consisting of the $\NP$'s f-term and a $\NOM$
case resource is embedded within a $\SUBJ$ attribute, which together
with the f-term associated with the $\VP$ yields the multiset
f-term associated with the $\S$.
(The interface between c-structure and f-terms is formalized
in \namecite{Johnson97} as a labelled deductive system).

\begin{eqnarray}
\hbox to 0.7in{\phon{Sandy}} & \NP & \sem{Sandy} : \NOM \limp e \label{e:she} \\
\hbox to 0.7in{\phon{snores}} & \VP & \lambda x . \sem{snores}(x) :
                \SUBJ\; e \limp t \label{e:snores}
\end{eqnarray}
\begin{equation} \label{e:SNPVP}
\S \;\longrightarrow\; 
  \begin{array}[t]{c} \NP\\ \SUBJ(\NOM, \down) \end{array} \; 
  \begin{array}[t]{c} \VP\\ \down \end{array} 
\end{equation}

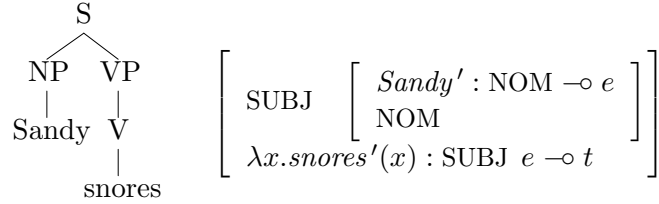
\begin{figure}
\begin{center}
\mbox{\lower3em\hbox{
\begin{picture}(54,78)(0,-78)
%
\put(24,-8){S}
\drawline(27,-12)(13,-22)
\put(6,-30){NP}
\drawline(13,-34)(13,-44)
\put(0,-52){Sandy}
\drawline(27,-12)(40,-22)
\put(33,-30){VP}
\drawline(40,-34)(40,-44)
\put(36,-52){V}
\drawline(40,-56)(40,-66)
\put(27,-74){snores}
\end{picture}}}
\hspace{0.2in}
\fs{ \SUBJ & \tfs{ 
                   \noatt{$\sem{Sandy} : \NOM  \limp e$} \\
                   \noatt{\NOM} } \\
     \noatt{$\lambda x . \sem{snores}(x) : \SUBJ\; e \limp t$}  }
\end{center}
\caption{\label{f:cf2} 
  The c-structure and f-term for \phon{She snores} generated
  by the fragment (\ref{e:she}--\ref{e:SNPVP}).  The f-term simplifies
  straightforwardly to type $t$, yielding the semantic labelling
  $\sem{snores}(\sem{Sandy})$. }
\end{figure}

This fragment generates the c-structure and f-term depicted in
Figure~\ref{f:cf2}.  The f-term simplifies to type $t$ in the following
steps:
\[
\infer{ \sem{snores}(\sem{Sandy}) : t }
{ \infer{ \sem{Sandy} : \SUBJ e }
  { \infer{ \sem{Sandy} : \SUBJ \NOM \limp \SUBJ e }
    { \sem{Sandy} : \SUBJ ( \NOM \limp e ) }
  & \SUBJ \NOM }
& \lambda x . \sem{snores}(x) : \SUBJ e \limp t 
}
\]

\subsection{Icelandic Quirky Case Marking} \label{s:Icelandic}

\def\drengurinn{\gp{\em drengurinn}{the-boy.\mbox{nom}}}
\def\drengina{\gp{\em drengina}{the-boys.\mbox{acc}}}
\def\kyssti{\gp{\em kyssti}{kissed}}
\def\vantar{\gp{\em vantar}{lacks}}
\def\stulkuna{\gp{\em st\'{u}lkuna}{the-girl.\mbox{acc}}}
\def\mat{\gp{\em mat}{food.\mbox{acc}}}
\def\hann{\gp{\em hann}{he.\mbox{nom}}}
\def\virdist{\gp{\em vir{\eth}ist}{seems}}
\def\elska{\gp{\em elska}{love}}
\def\hana{\gp{\em hana}{her.\mbox{acc}}}
\def\peninga{\gp{\em peninga}{money.\mbox{acc}}}
\def\vanta{\gp{\em vanta}{lack}}

\noindent
Quirky Case marking in Icelandic presents a more complex
array of linguistic data which exercises a wider range of f-term
machinery.  
This construction has proven difficult to encode in 
unification-based grammars, and has motivated several
non-monotonic extensions to the basic unification grammar
machinery, such as LFG's constraint equations and a complex
inheritance system in HPSG \cite{Sag:HPSG4}.\footnote{
As far as I am aware, the only feature structure account of Icelandic Quirky
Case Marking that does not make use of non-monotonic devices was given
by \namecite{Sag92}.  That account requires each NP to be associated
with {\em two} case features, which are threaded as a difference list
through the tree.  It would be interesting to investigate whether other
examples which motivate non-monotonic devices can be expressed using
purely monotonic constraints in this manner.}
The analysis presented here
demonstrates how the resource sensitivity of R-LFG provides a
simple way to encode the LFG analysis of \namecite{Andrews82}
without requiring recourse to complex extensions to the basic
machinery of R-LFG.

In Icelandic, subject $\NP$s are usually case marked nominative,
as in (\ref{e:icenorm}).  However, a few verbs, such
as \phon{vantar} `lacks' exceptionally case mark their
subject $\NP$s with accusative or some other non-nominative ``quirky''
case (\ref{e:icequirky}).  The subjects of subject raising verbs, such
as \phon{vir{\eth}ist} `seems', usually appear in nominative
case (\ref{e:iceraising}), but if the embedded verb is 
a quirky case assigning verb then the matrix subject is
assigned the quirky case, rather than nominative (\ref{e:iceqraising}).

\eenumsentence{
\item \mbox{\drengurinn\ \kyssti\ \stulkuna} \  \label{e:icenorm}\\
      \strut `The boy kissed the girl'
\item \mbox{\drengina\ \vantar\ \mat} \   \label{e:icequirky}\\
      \strut `The boys lack food'
\item \mbox{\hann\ \virdist\ \elska\ \hana } \ \label{e:iceraising}\\
      \strut `He seems to love her'
\item \mbox{\hana\ \virdist\ \vanta\ \peninga}\ \label{e:iceqraising}\\
      \strut `She seems to lack money'
}

This pattern of data receives a straightforward informal account
in terms of case assignment if we make the following assumptions:
\begin{itemize}
\item All NPs must receive exactly one case,
\item Quirky case marking verbs always assign a quirky case,
\item Case is preserved in Raising and other grammatical operations, and
\item Structural nominative case is only optionally assigned.
\end{itemize}

Thus if a subject $\NP$ receives a quirky case, then that must
be the case that it appears in.  On the other hand, if the
subject $\NP$ is not assigned a quirky case, then the only case
available is structural nominative case.

This account can be formalized in R-LFG as follows.  The
phrase structure rules for this Icelandic fragment are the following.
\begin{equation} \label{e:iceSNPVP}
 \S \;\longrightarrow\;
   \tcarray{ \NP \\ \SUBJ(\opt{\NOM}, \down) }
   \tcarray{ \VP \\ \down }
\end{equation}
\begin{equation} \label{e:iceVP}
 \VP \;\longrightarrow\;
   \tcarray{ \V \\ \down }
   \left( \tcarray{ \NP \\ \OBJ(\opt{\ACC},\down) } \right)
   \left( \tcarray{ \VP \\ \XCOMP\;\down } \right)
\end{equation}
The phrase structure rule (\ref{e:iceSNPVP}) differs from the
corresponding English rule (\ref{e:SNPVP}) in that it optionally
embeds a $\NOM$ case under the $\SUBJ$ attribute.  The phrase
structure rule (\ref{e:iceVP}) introduces a verb, an optional direct
object $\NP$ with optional accusative case marking,
and an optional $\VP$.  It embedds the direct object
$\NP$'s f-term under the $\OBJ$ attribute and the $\VP$'s f-term under
the $\XCOMP$ attribute, as is standard in LFG.

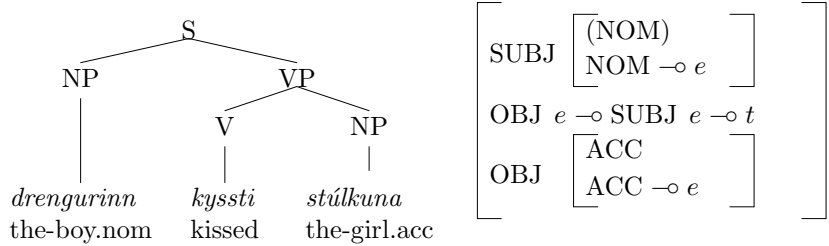
\begin{figure}
\def\NP{\mbox{$e$}}
\def\S{\mbox{$t$}}
\begin{center}\setlength{\unitlength}{0.00083333in}
\begingroup\makeatletter\ifx\SetFigFont\undefined%
\gdef\SetFigFont#1#2#3#4#5{%
  \reset@font\fontsize{#1}{#2pt}%
  \fontfamily{#3}\fontseries{#4}\fontshape{#5}%
  \selectfont}%
\fi\endgroup%
{\renewcommand{\dashlinestretch}{30}
\begin{picture}(5319,1389)(0,-10)
\path(3595,1287)(3445,1287)(3445,837)(3595,837)
\path(4420,1287)(4570,1287)(4570,837)(4420,837)
\path(3595,537)(3445,537)(3445,87)(3595,87)
\path(4420,537)(4570,537)(4570,87)(4420,87)
\path(370,762)(370,237)
\path(1270,462)(1270,237)
\path(2170,462)(2170,312)
\path(1270,687)(1720,837)(2170,687)
\path(370,987)(1045,1137)(1720,987)
\path(2995,1362)(2845,1362)(2845,12)(2995,12)
\path(4870,1362)(5020,1362)(5020,12)(4870,12)
\put(370,87){\makebox(0,0)[b]{\smash{{{\SetFigFont{10}{12.0}{\familydefault}{\mddefault}{\updefault}\drengurinn}}}}}
\put(1270,87){\makebox(0,0)[b]{\smash{{{\SetFigFont{10}{12.0}{\familydefault}{\mddefault}{\updefault}\kyssti}}}}}
\put(2170,87){\makebox(0,0)[b]{\smash{{{\SetFigFont{10}{12.0}{\familydefault}{\mddefault}{\updefault}\stulkuna}}}}}
\put(1270,537){\makebox(0,0)[b]{\smash{{{\SetFigFont{10}{12.0}{\familydefault}{\mddefault}{\updefault}V}}}}}
\put(2170,537){\makebox(0,0)[b]{\smash{{{\SetFigFont{10}{12.0}{\familydefault}{\mddefault}{\updefault}NP}}}}}
\put(1720,837){\makebox(0,0)[b]{\smash{{{\SetFigFont{10}{12.0}{\familydefault}{\mddefault}{\updefault}VP}}}}}
\put(370,837){\makebox(0,0)[b]{\smash{{{\SetFigFont{10}{12.0}{\familydefault}{\mddefault}{\updefault}NP}}}}}
\put(1045,1137){\makebox(0,0)[b]{\smash{{{\SetFigFont{10}{12.0}{\familydefault}{\mddefault}{\updefault}S}}}}}
\put(2920,987){\makebox(0,0)[lb]{\smash{{{\SetFigFont{10}{12.0}{\familydefault}{\mddefault}{\updefault}$\SUBJ$}}}}}
\put(3520,1137){\makebox(0,0)[lb]{\smash{{{\SetFigFont{10}{12.0}{\familydefault}{\mddefault}{\updefault}$\opt{\NOM}$}}}}}
\put(3520,912){\makebox(0,0)[lb]{\smash{{{\SetFigFont{10}{12.0}{\familydefault}{\mddefault}{\updefault}$\NOM\limp\NP$}}}}}
\put(3520,387){\makebox(0,0)[lb]{\smash{{{\SetFigFont{10}{12.0}{\familydefault}{\mddefault}{\updefault}$\ACC$}}}}}
\put(3520,162){\makebox(0,0)[lb]{\smash{{{\SetFigFont{10}{12.0}{\familydefault}{\mddefault}{\updefault}$\ACC\limp\NP$}}}}}
\put(2920,237){\makebox(0,0)[lb]{\smash{{{\SetFigFont{10}{12.0}{\familydefault}{\mddefault}{\updefault}$\OBJ$}}}}}
\put(2920,612){\makebox(0,0)[lb]{\smash{{{\SetFigFont{10}{12.0}{\familydefault}{\mddefault}{\updefault}$\OBJ\;\NP\limp\SUBJ\;\NP\limp\S$}}}}}
\end{picture}
}\end{center}
\caption{\label{f:ice-simple}
  The c-structure and f-term for the single clause non-quirky Icelandic example (\ref{e:icenorm}) generated by (\ref{e:iceSNPVP}--\ref{e:ices1c}).}
\end{figure}

The lexical entries (\ref{e:ices1a}--\ref{e:ices1c}) are required to generate 
the non-quirky single clause example (\ref{e:icenorm}).  The c-structure
and f-term associated with this example are shown in Figure~\ref{f:ice-simple}.
It is straightforward to check that this f-term reduces to $t$, labelled
with $\sem{kissed}(\sem{boy},\sem{girl})$.

\begin{eqnarray}
\hbox to 0.7in{\mbox{\em drengurinn}} & \NP & \sem{boy} : \NOM\limp e \label{e:ices1a}\\ 
\hbox to 0.7in{\mbox{\em st\'{u}lkuna}} & \NP & \sem{girl} : \ACC \limp e \\ 
\hbox to 0.7in{\mbox{\em kyssti}} & \V & 
 \lambda y \, \lambda x . \sem{kissed}(x,y) : \OBJ\; e \limp \SUBJ\; e \limp t 
\label{e:ices1c}
\end{eqnarray}

The single clause quirky case marked example is only slightly more
complex.  It can be described with the three additional lexical entries
(\ref{e:ices2a}--\ref{e:ices2b}).

\begin{eqnarray}
\hbox to 0.6in{\mbox{\em drengina}} & \NP & \sem{boys} : \ACC \limp e \label{e:ices2a} \\
\hbox to 0.6in{\mbox{\em mat}} & \NP & \sem{food} : \ACC \limp e \label{e:ices2a1} \\
\hbox to 0.6in{\mbox{\em vantar}} & \V & 
 \begin{array}[t]{l}
 \lambda y \, \lambda x . \sem{lacks}(x,y) : \OBJ\; e \limp \SUBJ\; e \limp t, \\
 \OBJ\;\ACC, \ul{\SUBJ\;\ACC} 
 \end{array} \label{e:ices2b}
\end{eqnarray}

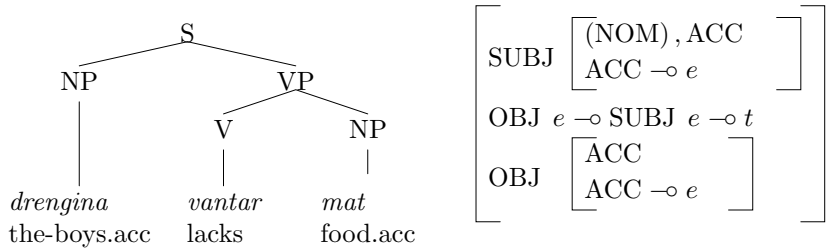
\begin{figure}
\def\NP{\mbox{$e$}}
\def\S{\mbox{$t$}}
\begin{center}\setlength{\unitlength}{0.00083333in}
\begingroup\makeatletter\ifx\SetFigFont\undefined%
\gdef\SetFigFont#1#2#3#4#5{%
  \reset@font\fontsize{#1}{#2pt}%
  \fontfamily{#3}\fontseries{#4}\fontshape{#5}%
  \selectfont}%
\fi\endgroup%
{\renewcommand{\dashlinestretch}{30}
\begin{picture}(5242,1389)(0,-10)
\path(3518,537)(3368,537)(3368,87)(3518,87)
\path(4343,537)(4493,537)(4493,87)(4343,87)
\path(293,762)(293,237)
\path(1193,462)(1193,237)
\path(2093,462)(2093,312)
\path(1193,687)(1643,837)(2093,687)
\path(293,987)(968,1137)(1643,987)
\path(2918,1362)(2768,1362)(2768,12)(2918,12)
\path(4793,1362)(4943,1362)(4943,12)(4793,12)
\path(3518,1287)(3368,1287)(3368,837)(3518,837)
\path(4643,1287)(4793,1287)(4793,837)(4643,837)
\put(293,87){\makebox(0,0)[b]{\smash{{{\SetFigFont{10}{12.0}{\familydefault}{\mddefault}{\updefault}\drengina}}}}}
\put(1193,87){\makebox(0,0)[b]{\smash{{{\SetFigFont{10}{12.0}{\familydefault}{\mddefault}{\updefault}\vantar}}}}}
\put(2093,87){\makebox(0,0)[b]{\smash{{{\SetFigFont{10}{12.0}{\familydefault}{\mddefault}{\updefault}\mat}}}}}
\put(1193,537){\makebox(0,0)[b]{\smash{{{\SetFigFont{10}{12.0}{\familydefault}{\mddefault}{\updefault}V}}}}}
\put(2093,537){\makebox(0,0)[b]{\smash{{{\SetFigFont{10}{12.0}{\familydefault}{\mddefault}{\updefault}NP}}}}}
\put(1643,837){\makebox(0,0)[b]{\smash{{{\SetFigFont{10}{12.0}{\familydefault}{\mddefault}{\updefault}VP}}}}}
\put(293,837){\makebox(0,0)[b]{\smash{{{\SetFigFont{10}{12.0}{\familydefault}{\mddefault}{\updefault}NP}}}}}
\put(968,1137){\makebox(0,0)[b]{\smash{{{\SetFigFont{10}{12.0}{\familydefault}{\mddefault}{\updefault}S}}}}}
\put(2843,987){\makebox(0,0)[lb]{\smash{{{\SetFigFont{10}{12.0}{\familydefault}{\mddefault}{\updefault}$\SUBJ$}}}}}
\put(3443,1137){\makebox(0,0)[lb]{\smash{{{\SetFigFont{10}{12.0}{\familydefault}{\mddefault}{\updefault}$\opt{\NOM},\ACC$}}}}}
\put(3443,912){\makebox(0,0)[lb]{\smash{{{\SetFigFont{10}{12.0}{\familydefault}{\mddefault}{\updefault}$\ACC\limp\NP$}}}}}
\put(3443,387){\makebox(0,0)[lb]{\smash{{{\SetFigFont{10}{12.0}{\familydefault}{\mddefault}{\updefault}$\ACC$}}}}}
\put(3443,162){\makebox(0,0)[lb]{\smash{{{\SetFigFont{10}{12.0}{\familydefault}{\mddefault}{\updefault}$\ACC\limp\NP$}}}}}
\put(2843,237){\makebox(0,0)[lb]{\smash{{{\SetFigFont{10}{12.0}{\familydefault}{\mddefault}{\updefault}$\OBJ$}}}}}
\put(2843,612){\makebox(0,0)[lb]{\smash{{{\SetFigFont{10}{12.0}{\familydefault}{\mddefault}{\updefault}$\OBJ\;\NP\limp\SUBJ\;\NP\limp\S$}}}}}
\end{picture}
}\end{center}
\caption{\label{f:ice-qsimple}
  The c-structure and f-term for the single clause quirky case
  example (\ref{e:icequirky}) generated 
  by (\ref{e:iceSNPVP}--\ref{e:ices2b}).}
\end{figure}

The lexical entry for the quirky case marking verb \phon{vantar}
`lacks' in (\ref{e:ices2b}) differs from that for the non-quirky verb
\phon{kyssti} `kissed' in that it assigns an accusative case to its
subject (in the underlined part of the f-term)
as well as to its object.  The c-structure and f-term for
(\ref{e:icequirky}) are depicted in Figure~\ref{f:ice-qsimple}.
Again, it is straightforward to check that the f-term reduces to $t$,
and is labelled with the $\lambda$-term $\sem{lacks}(\sem{boys},\sem{food})$.
Note that if the subject were replaced with a nominative $\NP$ the
f-term would no longer reduce to $t$, since the $\ACC$ case feature
embedded under the $\SUBJ$ attribute could not be consumed. 

\begin{figure}
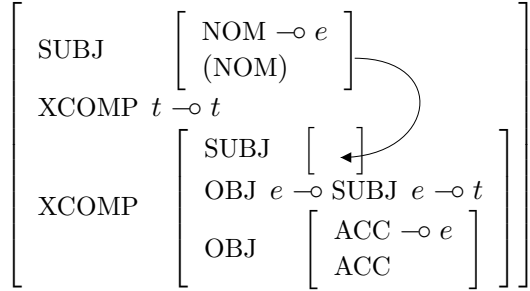

\begin{center}
\fs{ \SUBJ & \tfs{ \noatt{$\NOM\limp e$} \\
                  \noatt{$\opt{\NOM}$} } \nodepoint{a} \\
     \noatt{$\XCOMP\; t \limp t$} \\
     \XCOMP & \bfs{ \SUBJ & \fs{\ \nodepoint{b}\ } \\
                   \noatt{$\OBJ\; e \limp \SUBJ\; e \limp t$} \\
		   \OBJ & \bfs{\noatt{$\ACC\limp e$} \\
                              \noatt{$\ACC$} } } }
\anodecurve[r]{a}[r]{b}{0.5in}
\end{center}
\caption{\label{f:iceraising}
  The f-term for the non-quirky Subject Raising
  example (\ref{e:iceraising}) generated 
  by (\ref{e:iceSNPVP}--\ref{e:ices}).}
\end{figure}

The formalization of the non-quirky case Subject Raising example
(\ref{e:iceraising}) is very similiar to the standard LFG account
of Subject Raising \cite{Bresnan82}.  
The lexical entry (\ref{e:icerv})
for the Raising verb \phon{vir{\eth}ist} `seems' contains the path
equation $\SUBJ = \XCOMP\;\SUBJ$ which permits resources embedded 
under the $\SUBJ$ attribute to be restructured under the $\XCOMP\;\SUBJ$
attributes.  In this example, a resource of type $e$ is lowered
into the embedded clause.  The f-term associated with this example
is depicted in Figure~\ref{f:iceraising}.  Here we ignore the complexities
of pronominal binding, and treat the pronouns simply as NPs that consume a
nominative or accusative case resource.  It is straightforward
to check that this reduces to $t$, and is labelled with the $\lambda$-term
$\sem{seems}(\sem{loves}(\sem{he},\sem{her}))$. 

\begin{eqnarray}
\hbox to 0.6in{\mbox{\em vir{\eth}ist}} & \V & 
 \begin{array}[t]{l}
 \lambda P . \sem{seems}(P) : \XCOMP\; t \limp t, \\
 \SUBJ = \XCOMP\;\SUBJ 
 \end{array} \label{e:icerv} \\
\hbox to 0.6in{\mbox{\em elska}} & \V & 
 \begin{array}[t]{l}
 \lambda y \, \lambda x . love(x,y) : \OBJ\; e \limp \SUBJ\; e \limp t, \\
 \OBJ\;\ACC 
 \end{array} \label{e:ices}
\end{eqnarray}

\begin{figure}
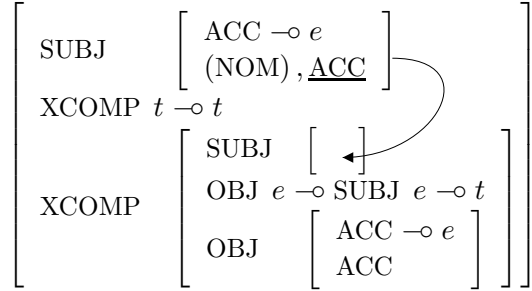

\begin{center}
\fs{ \SUBJ & \tfs{ \noatt{$\ACC\limp e$} \\
                  \noatt{$\opt{\NOM},\ul{\ACC}$} } \nodepoint{a} \\
     \noatt{$\XCOMP\; t \limp t$} \\
     \XCOMP & \bfs{ \SUBJ & \fs{\ \nodepoint{b}\ } \\
                   \noatt{$\OBJ\; e \limp \SUBJ\; e \limp t$} \\
		   \OBJ & \bfs{\noatt{$\ACC\limp e$} \\
                              \noatt{$\ACC$} } } }
\anodecurve[r]{a}[r]{b}{0.5in}
\end{center}
\caption{\label{f:iceqraising}
  The f-term for the quirky case marked Subject Raising
  example (\ref{e:iceqraising}) generated 
  by (\ref{e:iceSNPVP}--\ref{e:ices}).}
\end{figure}

The syntactic rules and lexical entries introduced above
that are independently needed to account for quirky case marking
in single clause constructions and for Subject Raising without
quirky case also correctly account for the interaction of those two
constructions, which was presented in (\ref{e:iceqraising}) on
page~\pageref{e:iceqraising}.  The f-term for this
example is shown in Figure~\ref{f:iceqraising}.

Just as in the single clause quirky case marking example (\ref{e:icequirky}),
the subject $\NP$ is assigned both an accusative case and an optional
nominative case, so only an accusative subject $\NP$ can appear.
If a nominative subject were inserted in matrix subject position
it could consume the optional
nominative case resource, but the accusative case resource assigned
by the quirky verb to the subject would not be consumed, and so an
f-term of type $t$ could not be derived.
It is straight-forward to check that the f-term depicted in Figure~\ref{f:iceqraising} 
simplifies to $t$, and that it is labelled with the $\lambda$-term
$\sem{seems}(\sem{lack}(\sem{she},\sem{money}))$, correctly
providing the required semantic interpretation.

\section{Conclusion}

\noindent
This paper has introduced a simplified version of LFG called R-LFG
in which a single representation called an f-term plays the role
of both f-description and f-structure.  This unification dramatically
simplifies the architecture of R-LFG, as compared to LFG augmented
with the Linear Logic interpretation machinery. 

Semantic interpretation in R-LFG exploits a Curry-Howard correspondence,
so semantic interpretation is obtained as a by-product of the 
syntactic type well-formedness checking process, and does not
need to be described in terms of stipulative, independently
specified semantic rules.

LFG's f-structure well-formedness constraints are re-expressed in
terms of feature resource dependencies, which permits them to be
checked by the same mechanism that performs semantic
interpretation. It is not implausible that this can be done for many,
if not most, LFG analyses, as many standard LFG analyses already have
a resource oriented character, and it seems that the ``core'' LFG
analyses of Raising, Control, etc., can be straightforwardly
reexpressed in R-LFG.  Treatments of phenomena such as quantifier
scoping, which motivate much of the glue logic work, still remain
to be developed, but there seems to be no principled problem here.


\bibliography{mj}
\endgroup
\end{document}